\documentclass[a4paper, 9 pt]{ieeeconf} 
\usepackage{tikz}
\usepackage{caption}    % For customized captions
\usepackage{subcaption} % For grouping figures
\usepackage{float}    % Required for fixing figure placement
\usepackage{booktabs} % For better table formatting
\usepackage{siunitx}  % For scientific notation formatting and number alignment
\definecolor{SPRaccent}{RGB}{175,17,64}

\IEEEoverridecommandlockouts

\makeatletter
\let\NAT@parse\undefined
\makeatother

\newcommand*\linkcolours{SPRaccent}

\usepackage{times}
\usepackage{graphicx}
\usepackage{amssymb}
\usepackage{textcomp}
\usepackage{gensymb}
\usepackage{amsmath}
\usepackage{url}
\usepackage{xurl}

\PassOptionsToPackage{hyphens}{url}\usepackage{hyperref}
\hypersetup{
colorlinks,
linkcolor=\linkcolours,
citecolor=\linkcolours,
filecolor=\linkcolours,
urlcolor=\linkcolours}

\usepackage{algorithm}
\usepackage{algorithmic}

\usepackage[labelfont={bf},font=small]{caption}
\usepackage[none]{hyphenat}
\usepackage{lettrine}
\usepackage{mathtools, cuted}

\usepackage[noadjust, nobreak]{cite}

\usepackage{tabularx}
\usepackage{amsmath}

\usepackage{float}

\usepackage{pifont}

\newcolumntype{Y}{>{\centering\arraybackslash}X}

\usepackage[]{placeins}

\usepackage{placeins}

\usepackage{tikz}

\usepackage[framemethod=tikz]{mdframed}

\usepackage{afterpage}

\usepackage{stfloats}

\usepackage{atbegshi}
\newcommand{\handlethispage}{}
\newcommand{\discardpagesfromhere}{\let\handlethispage\AtBeginShipoutDiscard}
\newcommand{\keeppagesfromhere}{\let\handlethispage\relax}
\AtBeginShipout{\handlethispage}

\usepackage{comment}
\usepackage{setspace}
\usepackage{amsmath}
\usepackage[most]{tcolorbox}

\usepackage{tgheros}

\usepackage[T1]{fontenc}
\usepackage[acronym]{glossaries}
\usepackage{adjustbox}

\usepackage{setspace}

\title{
    \Huge
    \vspace{10mm}
    \begin{flushleft}
    \onehalfspacing
    \textbf{Data and AI governance: Promoting equity, ethics, and fairness in large language models}

    \singlespacing
    \vspace{-2 mm}
    \normalsize
    Alok Abhishek$^{1,*}$, Lisa Erickson$^{2,*}$, and Tushar Bandopadhyay$^{3,*}$ 
    \vspace{3 mm}
    \footnotesize

    Edited by Swapnil Kumar and Emma Courtney
    \end{flushleft}
}

\author{\vspace{-10mm}
\thanks{\hspace{-4mm} \color{black}\line(1,0){\columnwidth}}
\thanks{\scriptsize\hspace{-3mm}$^{*}$Independent Researcher\vspace{2mm}}
\thanks{\scriptsize\hspace{-3mm}$^{1}$Email: alok@alokabhishek.ai\vspace{2mm}}
\thanks{\scriptsize\hspace{-3mm}$^{2}$Email: lisa.erickson@alum.mit.edu \vspace{2mm}}
\thanks{\scriptsize\hspace{-3mm}$^{3}$Email: tushar@kronml.com\vspace{2mm}}
\thanks{\scriptsize\hspace{-3mm}$^{}$The authors declare no conflict of interest.\vspace{2mm}}
\thanks{\scriptsize\hspace{-3mm}$^{}$© 2025 The Author(s)}}

\usepackage{fancyhdr}
\usepackage{lastpage}
\fancypagestyle{firststyle}

\usepackage[utf8]{inputenc}
\usepackage[english]{babel}
\renewcommand{\baselinestretch}{1.1}

\let\labelindent\relax
\usepackage{enumitem}

\begin{document}

\maketitle

\begin{abstract}
\vspace{-5mm}
\begin{tcolorbox}[boxrule=1.5pt,colback=SPRaccent!100!white,sharp corners,colframe=SPRaccent!0!black]
\Large\color{white}
\noindent\hspace{24mm}\textbf{HIGHLIGHTS}
\normalsize
\singlespacing
\begin{itemize}[leftmargin=*,labelindent=0mm,labelsep=2mm]\color{white}
  \item Adoption of Generative Artificial Intelligence (GenAI) and Large Language Models (LLMs) is growing rapidly. Bloomberg estimates that Generative AI will become a \$1.3 trillion market by 2032~{\hypersetup{citecolor=white}\cite{bloomberg2032}}.
  \vspace{1mm}
  \item The European Union (E.U.)'s Data Governance Act (2020) and Ethics Guidelines for Trustworthy AI (2019) provide initial regulatory frameworks. Significant practical implementation challenges still remain as current data and AI governance models do not sufficiently address GenAI specific complexities, highlighting the need for frameworks that effectively balance ethical compliance with real world applicability.
  \vspace{1mm}
  \item LLMs exhibit biases across many dimensions, such as gender, race and ethnicity, socioeconomic status, culture, religion, sexual orientation, disability, age, geography, political ideology, and stereotypes. Thematic analysis of biases in LLM responses show stereotypes, global and cultural dynamics, socioeconomic, and demographic related bias are prevalent in high severity and high risk responses~{\hypersetup{citecolor=white}\cite{abhishek2025beatsbiasevaluationassessment}}.
  \vspace{1mm}
  \item A data and AI governance framework that would facilitate the development of strategies to improve fairness, equity, and ethical alignment within the operational deployment of LLMs is urgently needed.
\end{itemize}

 \end{tcolorbox}

\small
\vspace{-1mm}
\singlespacing
\bf\noindent 
In this paper, we cover approaches to systematically govern, assess and quantify bias across the complete life cycle of machine learning models, from initial development and validation to ongoing production monitoring and guardrail implementation. Building upon our foundational work on the Bias Evaluation and Assessment Test Suite (BEATS) for Large Language Models~\cite{abhishek2025beatsbiasevaluationassessment}, the authors share prevalent bias and fairness related gaps in Large Language Models (LLMs) and discuss data and AI governance framework to address Bias, Ethics, Fairness, and Factuality within LLMs. The data and AI governance approach discussed in this paper is suitable for practical, real-world applications, enabling rigorous benchmarking of LLMs prior to production deployment, facilitating continuous real-time evaluation, and proactively governing LLM generated responses. By implementing the data and AI governance across the life cycle of AI development, organizations can significantly enhance the safety and responsibility of their GenAI systems, effectively mitigating risks of discrimination and protecting against potential reputational or brand-related harm. Ultimately, through this article, we aim to contribute to advancement of the creation and deployment of socially responsible and ethically aligned generative artificial intelligence powered applications.

\end{abstract}

%%%%%%%%%%%%%%%%%%%%%%%%%%%%%%%%%%%%%%%%%%%%%%%%%%%%%%%%%%%%%%%%%%%%%%%%%%%%%%%%

% As you follow along here, you can infer which commands correspond to which types of text. For example, the \PARstart command creates a large letter T that shows the start of the article.
\PARstart{A}{\MakeLowercase{rtificial}} Intelligence (AI) is fundamentally reshaping the technological landscape, driven by rapid innovation, accelerated adoption, and significant financial commitments from industries worldwide. Global investment in AI technologies is projected to grow, reaching approximately \$632 billion by 2028~\cite{idc2028, spglobal2028}. Within this broader AI expansion, investment in Generative AI (GenAI) is predicted to reach \$26 billion by 2027~\cite{idc2027}. In the long term, the GenAI market valuation is projected to reach an estimated \$1.3 trillion by 2032~\cite{bloomberg2032, precedence2024}, growing at an average annual growth rate of 36\% ~\cite{forrester2030}, further emphasizing the profound long term impact and potential. These projections highlight the scale of financial investments and underscore the accelerated pace at which generative technologies are being integrated across many sectors, signaling a paradigm shift in our use and interaction with AI in our everyday lives. As AI technologies evolve and their applications widen, the importance of comprehensive data and AI governance becomes increasingly crucial to ensure responsible and ethical development and use of AI~\cite{norori2021biashealthai,Alhosani2024aiinnovationingov}. 

The successful ethical implementation of AI and GenAI applications, particularly in fields such as natural language processing, computer vision, and decision support systems, relies on the quality and governance of training data. The emergence of large language models (LLMs), which utilize expansive and diverse training data corpora but are prone to reflecting existing societal prejudices present in the training data, emphasizes the need for comprehensive data and AI governance frameworks~\cite{bolukbasi2016mancomputerprogrammerwoman}. Effective data and AI governance is essential to address the considerable risks associated with these models.

Effective regulatory frameworks are foundational to responsible development of generative AI and can help address legal, ethical, and societal impacts while facilitating innovation. Critical elements of these regulatory frameworks include comprehensive legal infrastructures, international collaboration, and adaptive governance~\cite{luna2024navigating}. Comprehensive legal frameworks should explicitly address intellectual property rights, accountability, transparency, and safety, ensuring alignment with human-centered principles~\cite{PatriciaGomes2021airegulationframework,li2023legal}. 

The current AI regulatory landscape focuses on AI implementation, data privacy, fairness, and transparency. Although these regulations are still evolving, they are already driving organizations across sectors, including financial services ~\cite{JosephLee2021airegulationinfsi,Darko2025airegulationfs}, healthcare providers~\cite{GOVTRMAIINHC}, legal services~\cite{kapoor2024promisespitfallsartificialintelligence,magesh2024hallucinationfreeassessingreliabilityleading}, and government institutions~\cite{Alhosani2024aiinnovationingov,Mellouli2024issuesonaiinpubsec,Yigitcanlar2023aiingov}, to navigate emerging compliance requirements. These requirements affect machine learning operations and broader data governance practices. Current AI regulations primarily focus on ensuring responsible AI development and deployment in human-impact applications. These regulations address (1) data privacy and security, (2) fairness and non-discrimination, (3) explainability, (4) transparency, and (5) environmental sustainability.

\begin{enumerate}[label=(\arabic*),listparindent=\parindent]
    \item Existing data privacy and data sovereignty regulations across jurisdictions impose constraints on AI deployment, particularly in relation to data collection, processing, transfer, localization, and retention. In the European Union, the General Data Protection Regulation (GDPR) (\hyperlink{leg_1}{1}) mandates stringent consent and purpose limitation requirements. In the United States, the California Consumer Privacy Act (CCPA) (\hyperlink{leg_2}{2}) enforces consumer rights to access, delete, and opt out of data sharing. Beyond these, many countries across the globe have enacted national frameworks to assert data sovereignty and protect citizen privacy. 
    \par For instance, China has implemented a comprehensive set of laws including the Cybersecurity Law (\hyperlink{leg_6}{6}), the Data Security Law (\hyperlink{leg_7}{7}), and the Personal Information Protection Law (PIPL) (\hyperlink{leg_8}{8}), which impose strict data localization and cross-border data transfer restrictions. Similarly, Russia's Federal Law on Personal Data (\hyperlink{leg_9}{9}) requires that personal data of Russian citizens be stored and processed within national borders. \par
    India's Digital Personal Data Protection Act (DPDPA) (\hyperlink{leg_5}{5}) and Reserve Bank of India’s data localization mandates (\hyperlink{leg_4}{4}) reflect growing emphasis on data sovereignty in South Asia. In Latin America, Brazil’s Lei Geral de Proteção de Dados (LGPD) (\hyperlink{leg_10}{10}) parallels GDPR in its scope and enforcement mechanisms. Australia’s Privacy Act (\hyperlink{leg_11}{11}) continues to evolve to address digital and AI era privacy concerns. \par
    In the Middle East, the United Arab Emirates has adopted multiple frameworks, including the Federal Decree Law No. 45 of 2021 (\hyperlink{leg_12}{12}), the Dubai International Financial Centre (DIFC) Data Protection Law No. 5 of 2020 (\hyperlink{leg_13}{13}), and the Abu Dhabi Global Market (ADGM) Data Protection Regulations (\hyperlink{leg_14}{14}). Other Gulf states such as Saudi Arabia (\hyperlink{leg_15}{15}), Qatar (\hyperlink{leg_16}{16,17}), Bahrain (\hyperlink{leg_18}{18}), Oman (\hyperlink{leg_19}{19}), and Kuwait (\hyperlink{leg_20}{20}) have introduced similar data protection laws that emphasize both individual rights and national data governance.\par
    Collectively, these regulations are increasingly influencing how firms design, develop, and deploy AI and data platforms. Compliance with these frameworks now necessitates a deeper alignment of AI lifecycle governance with jurisdiction specific data handling requirements, thus adding complexity to cross-border AI scalability and innovation.\vspace{2mm}
    
    \item  Fairness and non-discrimination regulations, including the Equal Credit Opportunity Act (\hyperlink{leg_3}{3}) and the more targeted proposed E.U. AI Act (\hyperlink{leg_22}{22,23}), have strict requirements that AI systems remain free of biases related to race, gender, or religion. The E.U. AI act seeks to impose strict rules for high-risk AI applications to make sure customers are treated fairly. Fairness requirements creates difficulties in adoption of Gen AI as, GenAI models are trained on common internet data which might reinforce societal biases, leading to unfair outcomes~\cite{bolukbasi2016mancomputerprogrammerwoman,abhishek2025beatsbiasevaluationassessment,parrish2022bbq}. \vspace{2mm}
    
    \item Explainability and transparency requirements, like those outlined by the Basel Committee on Banking Supervision~\cite{Goodhart2011} and Model Risk Management  (\hyperlink{leg_27}{27}), demand that AI models be interpretable, auditable, and fair. These create difficulties for generative AI models, which often function as a "black box," and make it difficult to explain outputs or meet interpretability requirements. \vspace{2mm}
    
    \item The European Union’s Data Governance Act (2022) (\hyperlink{leg_21}{2}) and the Ethics Guidelines for Trustworthy AI (2019) (\hyperlink{leg_23}{23}) have laid down preliminary regulatory frameworks. These regulations focus on increasing trust in data sharing, and development of lawful, ethical, and robust AI applications.\vspace{2mm}
    
    \item While regulations such as the European Energy Efficiency Directive (EED) (\hyperlink{leg_24}{24}), which mandates data centers and technology intensive enterprises to optimize and transparently report their energy consumption, and standards like International Organization (ISO) 14001 (Environmental Management)~\cite{ISO14001}, which promotes a systematic approach to managing environmental impacts, directly shape technology infrastructure operations and, by extension, AI operations, broader governance frameworks and policy guidelines such as Organization for Economic Co-operation and Development (OECD) AI Principles (\hyperlink{leg_25}{25}) and the E.U. AI Act (\hyperlink{leg_22}{22}) explicitly integrate sustainability into the frameworks. These AI specific guidelines encourage resource efficient training and inference processes, minimal environmental footprints, and transparent reporting of AI models' carbon emissions, thereby aligning AI development with broader environmental sustainability objectives.
\end{enumerate}

Geopolitical considerations and AI sovereignty are also emerging areas of regulatory focus, with potential impacts for AI regulation. Emerging regulations may also address geopolitical concerns. National security and international competition in AI technology are driving trade restrictions, data localization, and sovereignty laws that could reshape global AI development (\hyperlink{leg_26}{26}).

While these regulatory frameworks provide a foundation for the security and privacy of data and ethical and responsible AI development, translating these high-level governance guidelines into effective real-world operational governance practice remains challenging. A critical gap lies in addressing the complex and nuanced ethical dilemmas and biases prominent in LLMs, which are trained on extensive but generic internet datasets prone to perpetuating societal biases and inequities. Despite regulatory efforts toward fairness, transparency, and explainability, practical implementation often struggles to keep pace with rapidly evolving AI capabilities. The scale and complexity of contemporary LLMs magnify these challenges, requiring adaptive and proactive methods to detect, quantify, and mitigate bias and ethical shortcomings. Therefore, a closer examination of specific bias-related shortcomings in leading LLMs is essential to identify areas where governance frameworks must evolve, bridging the gap between regulation and real-world ethical AI performance.

\large
\noindent\textbf{Ethics and bias related shortcomings in leading large language models} 
\normalsize %Extra line is necessary

A key focus in the ongoing efforts to improve AI governance is understanding and mitigating bias in GenAI systems. Bolukbasi et al., in their 2016 paper~\textit{“Man is to computer programmer as woman is to homemaker? debiasing word embeddings“}, demonstrated that word embedding models trained on common internet data like Google News encode and even amplify gender stereotypes to a dangerous extent~\cite{bolukbasi2016mancomputerprogrammerwoman}. This work showed how statistical correlations in training data can reinforce harmful societal prejudices. This paper laid the foundation for bias detection and drove impetus to reduce bias in early stage Natural Language Processing (NLP) systems.

Parrish et al., in the paper~\textit{“BBQ: A Hand-Built Bias Benchmark for Question Answering“}, introduce the Bias Benchmark for Question Answering (BBQ), a dataset designed to evaluate how social biases manifest in the outputs of question-answering (QA) models~\cite{parrish2022bbq}. This study highlighted that Natural Language Processing (NLP) models often reproduce harmful stereotypes, leading to biased outputs. 

In our recently published ArXiv paper~\textit{“BEATS: Bias Evaluation and Assessment Test Suite for Large Language Models“}, we introduced the Bias Evaluation and Assessment Test Suite (BEATS), a novel framework for evaluating Bias, Ethics, Fairness, and Factuality in Large Language Models (LLMs)~\cite{abhishek2025beatsbiasevaluationassessment}. Utilizing this framework, we demonstrated that 37.65\% of outputs generated by industry leading large language models contained some form of bias. Furthermore, 33.7\% of responses have either a high or medium level of bias severity or potential impact, highlighting a substantial risk of using these models in critical decision making systems.

Biases in LLMs span numerous dimensions, including stereotypes, cultural dynamics, individual and community biases. Biases related to age and gender form a core theme of reoccurring biases in the output of LLMs. Disability accommodations, gender roles, religious beliefs, economic disparities, and cultural influences are also prevalent. Prominent themes included systemic barriers, intergenerational technological adaptation, and socioeconomic challenges, highlighting complex interactions among demographic, cultural, and economic factors influencing LLM bias.

Studies such as Slattery et al.~\cite{slattery2025airiskrepositorycomprehensive} and Weidinger et al.~\cite{weidinger2021ethicalsocialrisksharm} underscore that the deployment of AI across high-stakes domains such as healthcare, law, finance, and public infrastructure introduces substantial risks that are frequently underestimated and inadequately governed. These risks include misinformation, systemic bias, privacy violations, weaponization, and regulatory failures and can arise from both human actions and autonomous AI behavior. Large language models have shown the potential to cause significant ethical and social harm, including discrimination, toxicity, information leakage, and automation-driven socioeconomic disruptions~\cite{weidinger2021ethicalsocialrisksharm,abhishek2025beatsbiasevaluationassessment}. These harms have been observed in research studies as well as in the real world and tend to disproportionately affect marginalized communities due to biased training data and uneven model performance.  

These findings clearly demonstrate that large language models exhibit biases across many dimensions, such as gender, race and ethnicity, socioeconomic status, culture, religion, sexual orientation, disability, age, geography, political ideology, and stereotypes. Therefore, incorporating LLMs into critical decision making applications, especially in domains such as healthcare~\cite{norori2021biashealthai,elshawi2019hypertension}, legal services~\cite{ magesh2024hallucinationfreeassessingreliabilityleading, kapoor2024promisespitfallsartificialintelligence}, finance~\cite{Qureshi2024AIinFinancialServices, zhang2019fairnessassessmentartificialintelligence}, and governance~\cite{ Alhosani2024aiinnovationingov,Mellouli2024issuesonaiinpubsec}, raises major risks and ethical concerns due to these inherent biases, potentially exacerbating systemic inequities. Given the high stakes repercussions of such biases, it is imperative to establish a rigorous governance framework that leverages statistical methodologies to systematically assess, mitigate, and manage biases. Such a governance framework would facilitate the development of strategies to improve fairness, equity, and ethical alignment within the operational deployment of LLMs.

\large
\noindent\textbf{Overview of data and AI lifecycle}\hypertarget{Data_and_AI_lifecycle}{}
\normalsize %Extra line is necessary

Data governance in AI involves the systematic management of data assets throughout their lifecycle from acquisition and storage to processing, deployment, and decommissioning. The lifecycle of AI systems is a structured and iterative process rather than a linear pipeline, aligning with contemporary frameworks such as the CDAC (Collect, Design, Assess, and Consume) model proposed by Silva et al. and calls to revise traditional AI lifecycle models to accommodate evolving operational demands and ethical considerations by Haakman et al.~\cite{desilva2021artificialintelligencelifecycle,haakman2021ailifecyclemodelsneed}.

The machine learning lifecycle spans five interdependent phases. It begins with data collection and preparation, where high-quality datasets are acquired from diverse sources and undergo rigorous curation, validation, auditing, and preprocessing to uphold ethical standards, mitigate bias, and support feature engineering. In the model development phase, appropriate algorithm selection, baseline establishment, hyperparameter optimization, fairness assessments, and explainability techniques are employed to develop performant and trustworthy models.

Deployment introduces the need for secure, scalable infrastructure and often involves containerizing models, exposing them through inference APIs, and implementing governance controls and guardrails that ensure compliance with organizational and regulatory standards. Post-deployment, continuous monitoring and maintenance are essential to detect model drift, track performance, and ensure ongoing fairness and ethical alignment for embedded governance checkpoints.

Finally, iterative improvement and retirement processes are driven by user feedback and production data insights. Models are refined continuously or retired systematically, with secure archival of model artifacts to ensure auditability and reproducibility. This holistic perspective, supported by frameworks like CDAC, highlights the iterative, sociotechnical nature of AI system lifecycles and the importance of aligning them with evolving best practices in AI ethics, transparency, and accountability.
\vspace{3mm}

\large
\noindent\textbf{Governance across the data and AI lifecycle}
\normalsize %Extra line is necessary

% Figure 1
\begin{figure*}[htbp]
\centering
\includegraphics[width=\textwidth]{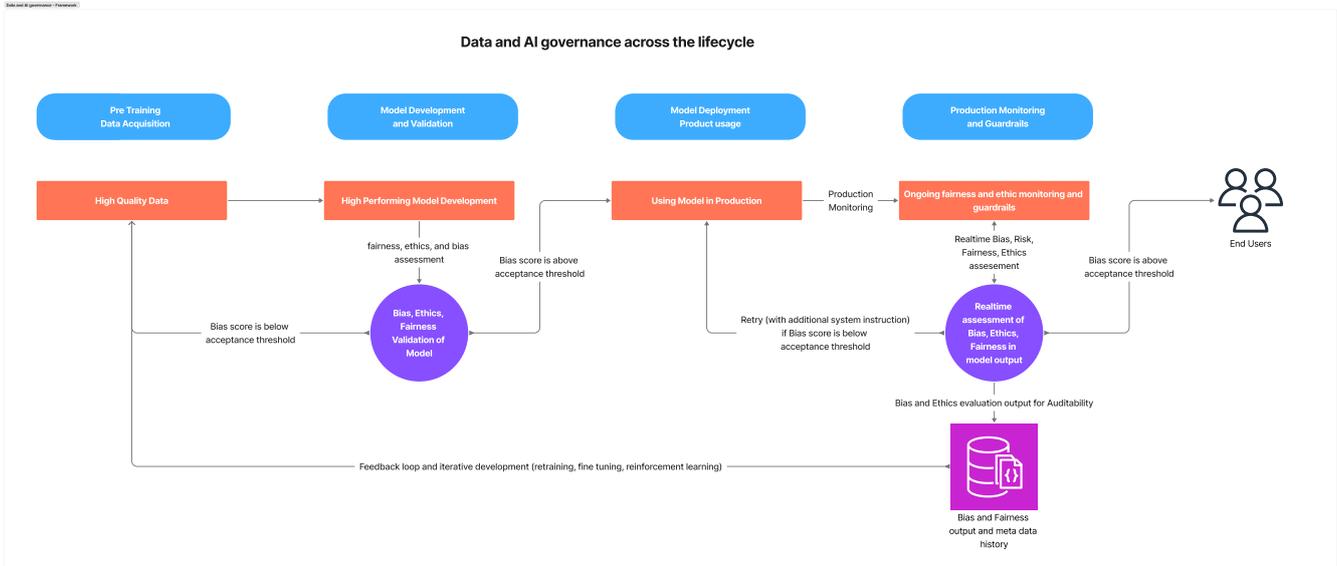}
\caption{System design of data and AI governance across the AI life cycle. The bias evaluation is performed as part of overall model evaluation before deploying the model in products and as an ongoing guardrail during model inference responses in production.}
\label{fig:bias_score_data_ai_gov_architecture}
\end{figure*}

A comprehensive data and AI governance approach to mitigating biases and ethical harms in generative AI models must systematically integrate governance practices across the entire data and machine learning (ML) lifecycle. 

It begins with the data collection and acquisition phase, where strategies include emphasizing source verification to ensure data diversity, reliability, and compliance with privacy regulations such as GDPR and CCPA. Additionally, regular demographic diversity and representativeness audits can be conducted, complemented by rigorous consent management and anonymization protocols for safeguarding sensitive information.

During the data preprocessing and labeling stage, bias detection techniques can be deployed to identify and correct systemic imbalances using statistical methods and normalization strategies. Transparent and standardized labeling protocols are implemented by diverse teams to minimize subjective and cultural biases. Comprehensive documentation, including meticulous version control and detailed data sheets, ensures data lineage traceability and accountability.

To ensure that AI systems are fair, transparent, and responsible, model development and training governance incorporate several critical practices. These include selecting fairness-aware algorithms that help minimize bias, conducting tests against misleading or adversarial inputs, and embedding ethical considerations throughout the design and development lifecycle. Independent ethics and bias review boards, along with formal fairness impact assessments, play a key role in maintaining accountability across the development process. Explainable AI (XAI) techniques, which make AI output easier to understand, are utilized to make model predictions more interpretable~\cite{wikipediaXAI}. Leading XAI methods, such as Shapley Additive Explanations (SHAP)~\cite{lundberg2017unifiedapproachinterpretingmodel}, Local Interpretable Model-Agnostic Explanations (LIME)~\cite{ribeiro2016whyitrustyou}, Partial Dependence Plots (PDPs)~\cite{moosbauer2022explaininghyperparameteroptimizationpartial}, and Counterfactual Explanations~\cite{verma2021counterfactualexplanationsmachinelearning}, help clarify how models arrive at specific results. These tools enhance transparency by documenting model assumptions, identifying potential limitations, and providing insights into the rationale behind predictions~\cite{elshawi2019hypertension}.

Validation and testing governance emphasize continual evaluation against established fairness metrics, employing cross group performance analysis to detect potential biases. Rigorous robustness and factual consistency tests, including counterfactual and out of sample evaluations, are performed alongside human in the loop verification processes to address edge cases and incorporate expert insights, fostering a feedback driven approach to bias mitigation. Human oversight remains crucial for accountability, emphasizing human in the loop models to ensure responsible AI operation. 

In the model deployment and monitoring phase, the framework maintains continuous fairness observability through real-time dashboards and routine audits. Ethical feedback mechanisms are strategically designed to prevent the reinforcement of biases during model updates. Incident response protocols provide structured mechanisms for addressing ethical breaches, ensuring timely resolution and user redress.

Finally, lifecycle governance emphasizes policies and practices that align with global AI regulations and ethical standards, supported by transparent stakeholder engagement, such as publicly accessible models and data cards. Continuous improvement initiatives integrate learning from ethical incidents, promoting an organizational culture deeply committed to ethical AI practices through regular training and external audits.

\large
\noindent\textbf{Application of the governance approach across the data and AI life cycle}
\normalsize %Extra line is necessary

Figure~\ref{fig:bias_score_data_ai_gov_architecture} shows a governance framework approach that covers the whole AI life cycle. This data and AI governance framework designed for real-world applications, enabling real-time assessment of LLM responses. This capability is essential for ensuring safe, responsible AI deployments that proactively mitigate discrimination risks and prevent potential brand or reputational damage.

This governance approach for data and AI is designed to integrates into two core phases of data and AI governance: model development and production guardrails. During the model development and validation phase, candidate models undergo evaluation for bias. Models achieving scores above the predefined acceptance threshold are approved for deployment into production environments. Conversely, models falling below this threshold trigger a retraining process, prompting a review of data curation practices to ensure training datasets do not reinforce societal inequities and are adequately representative of diverse global cultures.

In the production phase, this framework facilitates real-time assessments of model outputs, proactively identifying and mitigating ethical and fairness risks. Outputs exceeding acceptance thresholds proceed smoothly to subsequent stages, such as reaching the end users. However, if outputs falling below the acceptance thresholds the guardrails prevents the high risk responses from advancing further. The framework initiates a retry mechanism supplemented by additional prompt instructions, regenerating responses that align with ethical standards and fairness criteria.

This governance approach also incorporates an adaptive feedback mechanism, enabling continuous iterative improvements through retraining, fine-tuning, and reinforcement learning. This iterative cycle ensures that models learn from real world data insights and user feedback, facilitating continuous refinement and enhanced ethical alignment.

\large
\noindent\textbf{Limitations} 
\normalsize %Extra line is necessary

While the previously discussed data and AI governance framework provides comprehensive guidance for mitigating bias and ethical challenges across the lifecycle of LLM-powered applications, several limitations should be acknowledged:
\begin{enumerate}
    \item \textit{Dynamic regulatory and ethical landscape}: The governance framework described previously aligns with current regulatory guidelines and ethical considerations. However, the rapidly evolving global AI regulatory and fast evolving frontier LLM landscape implies that ongoing adjustments and refinements will be necessary. Governance approaches must remain flexible and adaptive to evolving regulatory, societal, and technological developments. \vspace{2mm}
    
    \item \textit{Scope and generalizability of the framework}: The framework discussed in this paper is primarily designed for governance in generative AI and LLM contexts. Although many governance practices can generalize to other AI domains, the specific methodologies and approaches may require adaptation for different types of AI systems, particularly those employing structured or multimodal data or those involving reinforcement learning or realtime interactive AI. \vspace{2mm}
    
    \item \textit{Limitations of bias measurement methods}: While the BEATS framework~\cite{abhishek2025beatsbiasevaluationassessment} provides structured evaluations of bias, fairness, ethics, and factuality, it uses LLMs as a judge paradigm where judge and generative models share similar training data, which is predominantly english and western culture centric. This could lead to a self-reinforcing mechanism, where a lack of global and diverse training data sets leads to a lack of sensitivity towards underrepresented or nondominant global viewpoints~\cite{ye2024justiceprejudicequantifyingbiases,zheng2023judgingllmasajudgemtbenchchatbot}. Therefore, there is a risk of evaluation scores representing fairness and ethical alignment, which are not global in nature.
\end{enumerate}

\large
\noindent\textbf{Conclusion} 
\normalsize %Extra line is necessary

Application of Generative AI, significantly accelerated by the introduction of transformer architecture by Vaswani et al.~\cite{vaswani2023attentionneed}, has fundamentally reshaped the landscape of applied artificial intelligence, influencing numerous industries ranging from finance and healthcare to governance and everyday applications worldwide. As these generative AI powered applications become deeply embedded in critical decision making processes, their societal implications have grown, raising the need for heightened scrutiny of ethical and regulatory compliance.

A growing body of scholarly research, combined with evolving regulatory frameworks, including the Equal Credit Opportunity Act (\hyperlink{leg_3}{3}), transparency and explainability mandates from the Basel Committee on Banking Supervision~\cite{Goodhart2011}, Model Risk Management guidelines (\hyperlink{leg_27}{27}), and fairness provisions proposed under the E.U. AI Act (\hyperlink{leg_22}{22,23}) has significantly contributed to development of fairer, more transparent, and accountable AI practices. Despite these regulatory advancements, research by Bolukbasi et al.~\cite{bolukbasi2016mancomputerprogrammerwoman}, Parrish et al.~\cite{parrish2022bbq}, and Abhishek et al.~\cite{abhishek2025beatsbiasevaluationassessment} clearly highlights persistent risks that Large Language Models (LLMs) perpetuate existing societal biases, potentially exacerbating systemic inequalities.

To bridge this gap between regulatory intent and practical application, we discuss a comprehensive data and AI governance framework. This governance approach systematically measures, monitors, and mitigates biases and ethical concerns, integrating fairness assessments, transparency, and privacy measures across the entire AI lifecycle. By proactively aligning operational practices with compliance requirements, including GDPR (\hyperlink{leg_1}{1}), the E.U. Data Governance Act (\hyperlink{leg_21}{2}), and sustainability oriented guidelines such as ISO 14001~\cite{ISO14001} and OECD AI Principles (\hyperlink{leg_25}{25}), organizations can enhance their AI systems' ethical alignment and regulatory adherence.

The adaptive nature of the discussed governance framework equips organizations to effectively navigate the dynamic regulatory and geopolitical landscapes, including data sovereignty and international AI governance concerns. Continuous monitoring and iterative enhancements embedded within this framework not only facilitate compliance but also foster stakeholder trust, mitigate reputational risks, and ensure the sustainable, equitable, and ethical deployment of AI technologies. This structured governance model advances the overarching goal of responsibly adopting the transformative potential of generative AI to create transparent, fair, and ethically aligned GenAI powered applications that genuinely benefit society.

\large
\noindent\textbf{Future research directions} 
\normalsize %Extra line is necessary

With the larger goal of contributing to the development of fairer LLMs that do not perpetuate societal biases and are suitable for use in critical decision making systems, we intend to continue future research in this area. To further extend the impact of the discussed governance framework, we outline several opportunities for future research:
\begin{enumerate}
    \item \textit{Empirical validation and framework optimization:} Empirical validation of the effectiveness of this governance framework across various industries would significantly strengthen the credibility and practical applicability of the approach. Detailed case studies and quantitative analyses will help refine the framework, identifying opportunities for optimization and tailored adaptations across diverse operational environments. \vspace{2mm}
    
    \item \textit{Exploration of multimodal GenAI governance:} Given the increasing prevalence of multimodal AI systems, future research to extend and refine the governance strategies that encompass multimodal data (e.g., images, video, and audio combined with text) will contribute greatly to the development of strategies for ethical and equitable GenAI applications. \vspace{2mm}
    
    \item \textit{Development of interactive governance tools:} We intend to develop practical, user-friendly tools and interfaces based on the Bias Evaluation and Assessment Test Suite (BEATS)~\cite{abhishek2025beatsbiasevaluationassessment}. These tools will facilitate real-time assessment, bias monitoring, and interpretability of model outputs, enabling practitioners and researchers to more readily identify, quantify, and mitigate biases in operational LLM powered GenAI applications.
\end{enumerate}

\large
\noindent\textbf{Acknowledgments}
\normalsize 

\noindent The authors extend sincere gratitude to the ~\textit{MIT Science Policy Review (SPR)} for the opportunity to contribute to this important discourse. We are especially thankful to ~\textit{Swapnil Kumar}, ~\textit{Emma Courtney}, and ~\textit{Audrey Bertin} for their support and insightful guidance throughout the editorial process. We also deeply appreciate the thoughtful and constructive feedback provided by the peer reviewers. 
\vspace{2mm}

\large
\noindent\textbf{Citation} %Space after is important
\normalsize

\noindent Abhishek, A., Erickson, L. \& Bandopadhyay, T. Data and AI governance: Promoting equity, ethics, and fairness in large language models. MIT Science Policy Review 6, 139-146 (2025). \url{https://doi.org/10.38105/spr.1sn574k4lp}.

\vspace{2mm}

\large
\noindent\textbf{Open Access} 

\noindent\includegraphics[width=2cm]{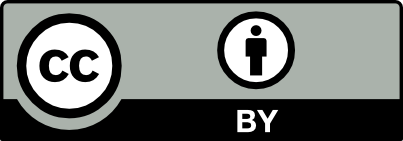}

\normalsize

\noindent This \textit{MIT Science Policy Review} article is licensed under a Creative Commons Attribution 4.0 International License, which permits use, sharing, adaptation, distribution and reproduction in any medium or format, as long as you give appropriate credit to the original author(s) and the source, provide a link to the Creative Commons license, and indicate if changes were made. The images or other third party material in this article are included in the article’s Creative Commons license, unless indicated otherwise in a credit line to the material. If material is not included in the article’s Creative Commons license and your intended use is not permitted by statutory regulation or exceeds the permitted use, you will need to obtain permission directly from the copyright holder. To view a copy of this license, visit \url{http://creativecommons.org/licenses/by/4.0/}.

\large
\noindent\textbf{Legislation cited} %Space after is important
\vspace{-2mm}
\normalsize

\footnotesize
\noindent

% sometimes you'll end up with weird spacing where your references section and legislation section don't line up properly in terms of their indents, i.e. the numbers are indented differently for legislation and references. This usually starts to happen if you have at least 10 pieces of legislation. You can fix this by modifying the "leftmargin" option in the line below. For example, you could change the line to something like: 
%\begin{enumerate}[label=(\arabic*),leftmargin=6.5mm]

\begin{enumerate}[label=(\arabic*),leftmargin=6.5mm]
    \item \hypertarget{leg_1}{European Union. General Data Protection Regulation (GDPR), Regulation (E.U.) 2016/679 (2016). Online: \url{https://eur-lex.europa.E.U./eli/reg/2016/679/oj}}

    \item \hypertarget{leg_2}{California State Legislature. California Consumer Privacy Act (CCPA) (2018). Online: \url{https://oag.ca.gov/privacy/ccpa}}

    \item \hypertarget{leg_3}{United States Congress. Equal Credit Opportunity Act (ECOA) (1974). Online: \url{https://www.consumerfinance.gov/rules-policy/regulations/1002/}}

    \item \hypertarget{leg_4}{Reserve Bank of India. Storage of Payment System Data (2018). Online: \url{https://www.rbi.org.in/Scripts/NotificationUser.aspx?Id=11244}}

    \item \hypertarget{leg_5}{Parliament of India. Digital Personal Data Protection Act (2023). Online: \url{https://www.meity.gov.in/static/uploads/2024/06/2bf1f0e9f04e6fb4f8fef35e82c42aa5.pdf}}

    \item \hypertarget{leg_6}{National People's Congress of China. Cybersecurity Law (2016). Online: \url{http://www.cac.gov.cn/2016-11/07/c_1119867116.htm}}

    \item \hypertarget{leg_7}{National People's Congress of China. Data Security Law (2021). Online: \url{http://www.npc.gov.cn/npc/c30834/202007/3b5d6944cbb74d3b9b8e5a4f6f3097e4.shtml}}

    \item \hypertarget{leg_8}{National People's Congress of China. Personal Information Protection Law (PIPL) (2021). Online: \url{http://www.npc.gov.cn/englishnpc/c23934/202111/2b1f0e1a2e3e4a2b9b8e5a4f6f3097e4.shtml}}

    \item \hypertarget{leg_9}{Russian Federation. Federal Law on Personal Data (2006). Online: \url{https://en.wikipedia.org/wiki/Data_protection_(privacy)_laws_in_Russia}}

    \item \hypertarget{leg_10}{Brazil. Lei Geral de Proteção de Dados Pessoais (LGPD) (2018). Online: \url{http://www.planalto.gov.br/ccivil_03/_ato2015-2018/2018/lei/l13709.htm}}

    \item \hypertarget{leg_11}{Parliament of Australia. Privacy Act 1988. Online: \url{https://www.legislation.gov.au/Details/C2023C00210}}

    \item \hypertarget{leg_12}{United Arab Emirates. Federal Decree-Law No. 45 of 2021 on the Protection of Personal Data. Online: \url{https://u.ae/en/about-the-uae/digital-uae/data/data-protection-laws}}

    \item \hypertarget{leg_13}{Dubai International Financial Centre. Data Protection Law No. 5 of 2020. Online: \url{https://www.difc.ae/business/laws-regulations/data-protection/}}

    \item \hypertarget{leg_14}{Abu Dhabi Global Market. Data Protection Regulations 2021. Online: \url{https://www.adgm.com/operating-in-adgm/doing-business/data-protection}}

    \item \hypertarget{leg_15}{Saudi Data and AI Authority. Personal Data Protection Law (PDPL) (2021). Online: \url{https://sdaia.gov.sa/en/Research/Pages/DataProtection.aspx}}

    \item \hypertarget{leg_16}{State of Qatar. Law No. 13 of 2016 on Personal Data Protection. Online: \url{https://www.motc.gov.qa/en/documents/document/law-no-13-2016-concerning-personal-data-privacy-protection}}

    \item \hypertarget{leg_17}{Qatar Financial Centre. Data Protection Regulations 2021. Online: \url{https://www.qfc.qa/en/operating-in-the-qfc/legal-and-regulatory/data-protection}}

    \item \hypertarget{leg_18}{Kingdom of Bahrain. Law No. 30 of 2018 Promulgating the Personal Data Protection Law. Online: \url{https://www.legalaffairs.gov.bh/Media/LegalPDF/K3001.pdf}}

    \item \hypertarget{leg_19}{Sultanate of Oman. Royal Decree No. 6/2022 Promulgating the Personal Data Protection Law. Online: \url{https://www.mola.gov.om/Download.aspx?Path=royal/2022-0006.pdf}}

    \item \hypertarget{leg_20}{State of Kuwait. Data Privacy Protection Regulation (2023). Online: \url{https://www.citra.gov.kw/sites/en/Pages/Data-Privacy-Protection.aspx}}

    \item \hypertarget{leg_21}{European Union. Data Governance Act (2022). Online: \url{https://digital-strategy.ec.europa.E.U./en/policies/data-governance-act}}

    \item \hypertarget{leg_22}{European Union. Artificial Intelligence Act (2024). Online: \url{https://artificialintelligenceact.E.U./}}

    \item \hypertarget{leg_23}{European Union. Ethics Guidelines for Trustworthy AI (2019). Online: \url{https://digital-strategy.ec.europa.E.U./en/library/ethics-guidelines-trustworthy-ai}}

    \item \hypertarget{leg_24}{European Union. Energy Efficiency Directive (2015). Online: \url{https://energy.ec.europa.E.U./topics/energy-efficiency/energy-efficiency-targets-directive-and-rules/energy-efficiency-directive_en}}

    \item \hypertarget{leg_25}{Organization for Economic Co-operation and Development (OECD). Recommendation of the Council on Artificial Intelligence (2019). Online: \url{https://oecd.ai/en/ai-principles}}

    \item \hypertarget{leg_26}{UK Government. International AI Safety Report (2025). Online: \url{https://www.gov.uk/government/publications/international-ai-safety-report-2025}}

    \item \hypertarget{leg_27}{United States. Office of the Comptroller of the Currency. Supervisory Guidance on Model Risk Management (2011). Online: \url{https://www.occ.gov/news-issuances/bulletins/2011/bulletin-2011-12.html}}
\end{enumerate}

\vspace{2mm}

\large
\noindent\textbf{References} 
\vspace{-12mm}
\normalsize
\bibliographystyle{naturemag}
\bibliography{bibliography}

\end{document}